\documentclass{article}
\usepackage[english]{babel}
\usepackage[textheight=9in,
    textwidth=16cm,
    top=1in,
    headheight=12pt,
    headsep=25pt,
    footskip=30pt]{geometry}
\usepackage{graphicx}
\usepackage[utf8]{inputenc} 
\usepackage[T1]{fontenc}    
\usepackage{hyperref}       
\usepackage{url}            
\usepackage{booktabs}       
\usepackage{amsfonts}       
\usepackage{nicefrac}       
\usepackage{microtype}      

\usepackage{textcomp}
\usepackage{subcaption}
\usepackage{wrapfig}

\usepackage{amsmath}
\usepackage{amsthm}
\usepackage{amssymb}
\usepackage{amsfonts}

\newcommand{\keywords}[1]{\textbf{Keywords: }#1}

\newcommand{\textdef}[1]{\text{#1}}
\newcommand{\N}{\mathbb{N}}
\newcommand{\R}{\mathbb{R}}

\renewcommand{\P}{\mathbb{P}}

\newcommand{\T}{\mathcal{T}}

\theoremstyle{definition}
\newtheorem{definition}{Definition}

\theoremstyle{plain}
\newtheorem{lemma}{Lemma}
\newtheorem{theorem}{Theorem}
\newtheorem{corollary}{Corollary}

\theoremstyle{remark}

\newtheorem{example}{Example}

\begin{document}
\title{Analysis of Drifting Features\thanks{We gratefully acknowledge funding by the BMBF under grant number 01 IS 18041 A.}}
%
%
%
%

\author{%
  Fabian Hinder \\
  Cognitive Interaction Technology (CITEC)\\
  Bielefeld University\\
  Inspiration 1, D-33619 Bielefeld, Germany \\
  \texttt{fhinder@techfak.uni-bielefeld.de} \\
  \\
  Jonathan Jakob \\
  Cognitive Interaction Technology (CITEC)\\
  Bielefeld University\\
  Inspiration 1, D-33619 Bielefeld, Germany \\
  \texttt{jjakob@techfak.uni-bielefeld.de} \\
  \\
  Barbara Hammer \\
  Cognitive Interaction Technology (CITEC)\\
  Bielefeld University\\
  Inspiration 1, D-33619 Bielefeld, Germany \\
  \texttt{bhammer@techfak.uni-bielefeld.de} \\
}

\maketitle              
\begin{abstract}
The notion of concept drift refers to the phenomenon that the distribution, which is underlying the observed data, changes over time. 
We are interested in an identification of those features, that 
are most relevant for the observed drift.
We distinguish between drift inducing features, for which the observed feature drift cannot be explained by any other feature, and faithfully drifting features, which correlate with the present drift of other features.
This notion gives rise to minimal subsets of the feature space, which are able to characterize the observed drift as a whole.
We relate this problem to the problems of feature selection and feature relevance learning, which allows us to derive a detection algorithm. We demonstrate its usefulness on different benchmarks.

\end{abstract}
\section{Introduction}
One  fundamental assumption in classical machine learning is the fact, that observed data are i.i.d.\ according to some unknown underlying probability measure $P_X$, i.e.\ the data generating process is stationary. 
Yet, this assumption is often violated 
in the context of real world problems: models are subject to seasonal changes, changed demands of individual costumers, ageing of sensors, etc. In such settings, 
life-long model adaptation rather than classical batch learning is required.
Since drift, i.e.\ the fact that data are no longer identically distributed, is a major issue in many real-world
applications of machine learning,
quite a few approaches address the question
of data modeling in the context of drift  \cite{DBLP:journals/cim/DitzlerRAP15}.

Depending on the application domain, different types of drift occur.
Covariate shift  refers to the  situation of  training and test set having different marginal distributions   \cite{5376}. Learning for data streams, which is often encountered as supervised learning scenarios, extends this setting to an unlimited (but usually  countable) stream of observed data \cite{asurveyonconceptdriftadaption}.
Thereby, one distinguishes virtual drift, i.e.\ non-stationarity of the marginal distribution only, and real drift, i.e.\ non-stationarity  of the posterior as well. 
A huge variety of different approaches exists,
including windowing techniques, active methods, or ensemble technologies as prominent examples    \cite{DBLP:journals/cim/DitzlerRAP15,asurveyonconceptdriftadaption,DBLP:journals/sigkdd/GomesRBBG19,sea,DBLP:journals/kais/LosingHW18}. 

Interestingly, a majority of approaches deals with supervised scenarios, aiming for a small interleaved train-test error; this is accompanied by  methods, which aim for a detection of drift or change points in given data sets \cite{Aminikhanghahi:2017:SMT:3086013.3086037,DBLP:journals/tnn/AlippiBR17},
a quantification of the specific type of drift \cite{DBLP:journals/kais/GoldenbergW19},
or a modeling of the drift process by dynamic models \cite{DBLP:journals/tnn/Roveri19}.
Further, first theoretical results accompany such technologies by provable bounds on
their generalization ability \cite{DBLP:journals/eswa/MelloVFB19}.
So far, very few approaches address the question of explaining the relevant factors which occur in a drift, including 
global feature selection  in data streams \cite{asurveyonfeaturdrift}, 
or an  identification of features with particularly prominent drift 
\cite{DBLP:journals/corr/WebbLPG17}. 

The purpose of our contribution is to introduce a feature wise notion of drift, that captures potential causal inferences between the given features. Unlike the approach \cite{DBLP:journals/corr/WebbLPG17}, we do not refer to drift of the marginal feature distributions, since this might be misleading in the sense that it does not capture cases of feature drift caused by changing feature correlations rather than changing marginal distribution of single features. Hence, we propose a
novel formalization, which is based on the rationale, that
underlying causal relations of some features might induce drift dependencies. Keeping this in mind, we are able to give  a definition and first algorithmic solutions for minimal explanations for observed drift.

This paper is organized in three main sections:
In the first, we examine the problem and give a definition for three different feature categories, capturing some type of causality for observed drift. Furthermore, we examine, under which circumstances these notions are distinct from each another, thereby establishing a proper setup of the problem. 
In the second section, we use our previous results, to compare the problem of drifting-feature-analysis to existing problems of feature relevance determination, which is a well-studied problem in static analysis for model explanation \cite{YuLiu,Rudnicki2019Chapter2A,InterpretationofLinearClassifiersbyMeansofFeatureRelevanceBounds}. From this comparison, we obtain a method for drifting-feature-analysis. This is evaluated in the last section. 
 Due to page limitations, all proofs, that are part of this contribution, are included in the supplement only.

\section{Drifting-Feature-Analysis Problem}
\label{sec:setup}

In the usual, time invariant setup of machine learning, one considers a generative process $\P_X$, i.e. a probability measure, on $\R^d$. In this context, one views the realizations of $\P_X$-distributed random variables $X_1,...,X_n$ as samples.
Depending on the objective, learning algorithms try to infer the data distribution based on these samples, or, in the supervised setting, the posterior distribution. We will not distinguish between these settings and only consider distributions in general, this way subsuming the notion of both, real drift and virtual drift.

Many processes in real-world applications are not time independent, so it is reasonable to incorporate time into our considerations. One prominent way to do so, is to consider an index set $\T$, representing time, and a collection of probability measures $p_t$ on $\R^d$, indexed over $\T$, which may change over time \cite{asurveyonconceptdriftadaption}. 
In the following, we investigate the relationship of those $p_t$, with concept drift referring to the property, that the probability measures $p_t$ vary with $t$. Notice, that we will usually use the shorthand notion drift instead of concept drift.

In this contribution, we are interested in determining, which features change over time and how this feature change can explain observed drift. Due to possible redundancies of features, we will distinguish 
between three categories: non drifting features, faithfully drifting features, and drift inducing features. \textit{Non drifting features } refer to features, for which the distribution  does not change over time; \textit{faithfully drifting features} identify features, for which  the distribution changes, but the observed change can be explained by the fact, that the features depend on other drifting features and they change according to this relationship; \textit{drift inducing features } refer to features, for which the distribution changes, and this change cannot be explained by the drift of other features. We will refer to this classification problem as \textit{drifting-feature-analysis (problem)}.
%
\begin{example}
\label{elp:physics}
Suppose, that we examine a closed cylinder, filled with gas or steam. We measure temperature and pressure. If we heat up the cylinder, then the pressure will rise as a consequence. Therefore, heating up the cylinder induces drift, i.e.\ temperature is a drift inducing feature, while pressure changes as a result, i.e.\ pressure is a faithfully drifting feature.
\end{example}

 Drifting-feature-analysis, as stated above, is not yet well defined.
 In the remainder of this section, we will provide two different mathematical definitions of
 this intuition and investigate their relationship.
We will first consider a definition, which relies on the behavior of single features, but there exist settings, where this 
can lead to unintuitive behavior. Hence, a more complex definition, which takes into account
sets of features, will be given. 
%
To start with, we will recap, how the drift of a distribution is defined. From there, we carve out the precise definitions of the above terms.

\subsection{Definition of Drift and Drifting Features}
\label{sec:def_drift}

Let us first fix the notation, which we will use for the rest of this paper. 
A stream is a potentially infinite sequence of sample points $x_1$, $x_2$, \ldots, $x_n$, \ldots, that origin from some unknown distribution, which may change over time. Hence, those samples are no longer necessarily i.i.d.\ realizations of an underlying random variable $X$. 
It can be shown \cite{driftingdata}, that this problem may be overcome by adding "time stamps" to our samples: the distribution of data $(x_1, t_1)$, $(x_2,t_2)$, \ldots, where $t_i$ represents the time associated with the arrival of $x_i$ (e.g. $t_i = i$) can be considered as i.i.d.\ realizations of a random variable $(X, T)$. We will later rely on this property to derive algorithms for the drifting-feature-analysis.

We are interested in the relevance of features for an observed drift, hence, we introduce a notation for the features of a data point $x_i$:
$X$ denotes the random variable that represents observed data, which are element of $\R^d$, for simplicity -- notice, that the theory can be easily extended to more general spaces.
We denote the features of $X$ in vector notation, i.e.\ $X = (X_1,...,X_d)$, where $X_i$ represents the $i$-th feature. For a set $S \subset \{1,...,d\}$ of indices $S=\{s_1,...,s_{|S|}\}$,  $X_S := (X_{s_1},...,X_{s_{|S|}})$ denotes the vector consisting of the features contained in $S$. 
We use the special notation $X_i := X_{\{i\}}$ in the case, where $S = \{i\}$ contains only a single feature and $R_i := \{i\}^C = \{1,...,i-1,i+1,...,d\}$ in the case, where we use all but one feature.


We denote the distributions of $X$ resp.\ $T$ by $\P_X$ resp.\ $\P_T$. Furthermore, we denote the conditional distribution of $X$ given $T$ by $\P_{X|T=t}(x) = p_t(X=x)$.
This allows us to directly adapt the definition of drift from \cite{asurveyonconceptdriftadaption} to our setup:

\begin{definition}
\label{def:drift}
The random variable $X$ has \textdef{drift} (or is drifting) iff
\begin{align}
    \P_T\times \P_T(\{(t,s)\in\T^2\:|\: p_t(X) &\neq p_s(X) \}) > 0.
\end{align}

We say, that the random variable $Y$ has \textdef{conditional drift} (or is conditionally drifting) given $X$ iff
\begin{align}
    \P_{T|X=x}\times\P_{T|X=x}(\{(t,s)\in\T^2\:|\: (p_t(Y|x) &\neq p_{s}(Y|x) \}) > 0
\end{align}
holds on a $\P_X$ non-null set.
\end{definition}

Hence, $X$ has drift, if and only if its distribution changes over time.
We can also focus on a distribution during a time-window, i.e.\ $\P_{X|T \in [t_0,t_1]}$, where $t_0,t_1$ represent the boundaries of the time window. 
Furthermore, we can  naturally  handle varying sample densities and potentially missing samples, that are due to phenomena like sensor failure for example. In addition to the case of discrete time, i.e. $\T \subset \N$, which is the most common case in the literature, the notion of drift presented above coincides perfectly with the notion of drift given by \cite{asurveyonconceptdriftadaption}.
Notice, that in the case of $Y$ being  the label and $X$  the features of a classification or regression problem, conditional drift is is also referred to as real drift \cite{asurveyonconceptdriftadaption}.

Now, that we defined drift of a process
on a set of features, we would like to define \emph{drift of a single feature}. A direct approach, also considered in the work \cite{DBLP:journals/corr/WebbLPG17}, is to consider feature-wise drift as can be observed in the marginal distribution of a single feature. However, this would miss relevant settings:
as a first example, consider a classification problem with real drift and stable class distributions, i.e.\ $p(y|X)$ changes but $p(y)$ does not. 
Here, the marginal distribution of $y$ does not change.  Still, we would 
 say, that $y$ has drift.
As a second example, consider two Gaussians in the unit square at positions $(-1,-1)$ and $(1,1)$, and at positions $(1,-1)$ and $(-1,1)$, respectively.  This  gives rise to two different joint distributions, but identical marginal distributions. 

Hence, it is apparent,  
that drift may happen in the way features interact, rather than in the marginal distribution of features themselves. Hence, we build the definition of feature drift on a the relation of 
a feature distribution to others. 
To facilitate this, we rely on conditional drift, which enables us to formalize that
a connection of features is changing. 
This gives rise to the following definition:
\begin{definition}
\label{def:featuredrift}
%
A feature $X_i$ is called \textdef{drifting}, iff there exists a set $R$, such that $X_i$ is conditionally drifting given $X_R$. It is called \textdef{strongly drifting} if we may choose $R = R_i$, otherwise it is called \textdef{weakly drifting}. 
\end{definition}

As a special case, we obtain the notion of marginal distributional change, i.e.\ drift of the value distribution of a single feature, by conditioning on the empty set. Yet the definition is more general since it allows us to capture changes of the relation of a  feature  to others.  Depending on the specific conditioning,
we can distinguish different types of drift.
Per definition, for any weakly drifting but not strongly drifting feature $X_i$, 
we can find a set of features $S$ such that $X_i$ is not drifting given $X_S$. 
This circumstance may be interpreted as the fact, that the  relation between $X_i$ and $X_S$ does not change over time, so that an observed drift is induced by the drift in $X_S$ rather than a drift of $X_i$ itself, i.e.\ one could attribute causality of the observed drift to $X_S$ rather than $X_i$.
Yet, this argument also demonstrates that this definition of drift is not monotonic w.r.t.\  the set of features used for conditioning. This means, that if $X_i$ is conditionally drifting given $X_S$, then both adding and removing a feature to resp.\ from $S$ may cause, that $X_i$ does no longer drift conditionally. 
This shows, that this definition of feature wise drift may serve as a starting point but most probably requires further refinement, in particular to allow intuitive interpretation and its efficient algorithmic computation.
Hence we will give another definition of feature wise drift in the next section, which is based on ideas of causal explanations. 
It will turn out, that this notion of drift is better behaved, and matches suitably well with the definition of feature wise drift given above.

\begin{figure}[!tb]
    \centering
      \includegraphics[width=10cm]{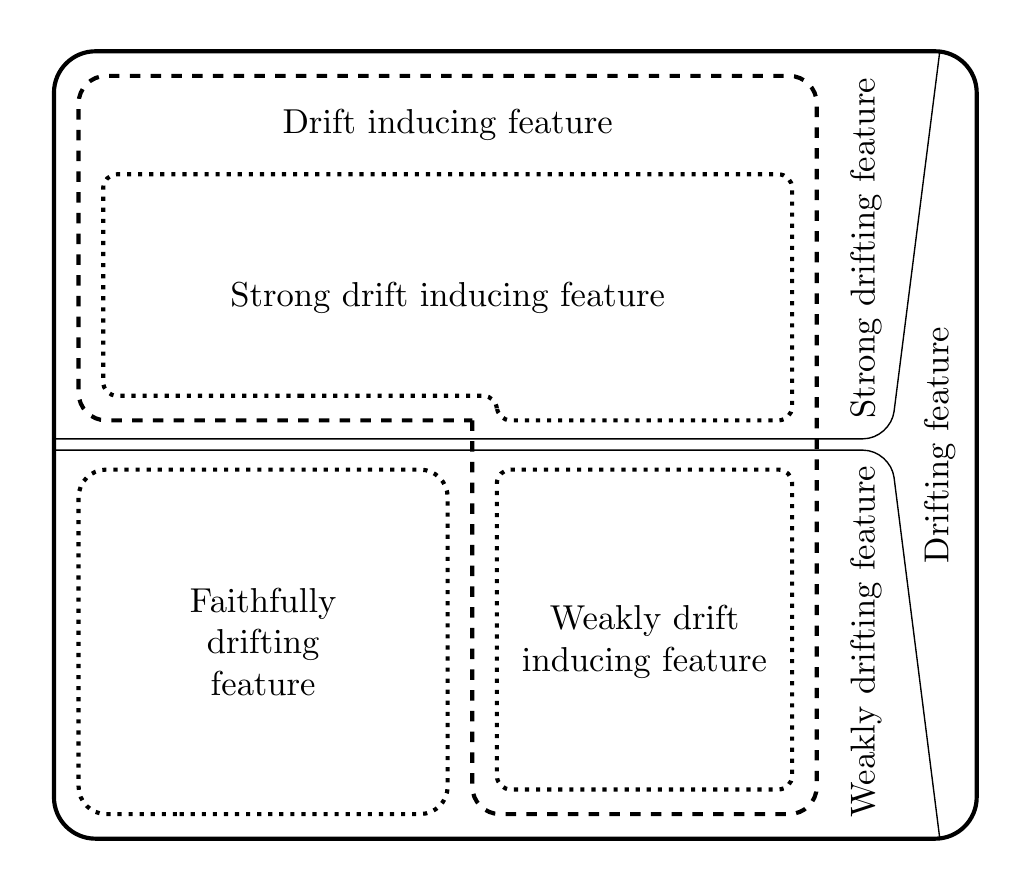}
    \caption{Visualization of definition~\ref{def:featuredrift}, \ref{def:drift_inducing:feature} and \ref{def:faithfully}, theorem~\ref{thm:inducing_and_drifting} and corollary~\ref{thm:weakly_inducing_is_weakly_drifting}.}
    \label{fig:visualization_def}
\end{figure}

\subsection{Drift Inducing Features}
So far, we have considered the problem of feature wise drift. We will now consider \emph{sets of features}, that are able to explain observed drift, and derive subsequent notions for feature wise drift afterwards. The latter are related to the above definition of strongly and weakly drifting features, but they are not identical. In section~\ref{sec:well_suited}, we will formalize a criterion, that allows us to identify the relevant groups of features, and gives rise to an exact formulation of the drifting-feature-analysis problem. The different definitions of types of drifting features and their relations, which we will consider, are visualized in figure~\ref{fig:visualization_def}.

As a motivation, let us reconsider example~\ref{elp:physics}: In practice  we expect measurements to be coupled, to some point, by the laws of physics. Some features may change over time, but we do not expect the coupling laws  to change. Instead, if we manipulate some of their features, coupled  features  change accordingly. In this setup,  the change of dependent features is induced by the change of manipulated  features  -- we call the latter the inducing features. Clearly, the drift of the inducing features depends on time (we start to manipulate them at some point in time after all), but the drift of the other features depends on the inducing features and not on time itself. 
This idea yields the following definition of
features sets which are capable of inducing the observed drift:

\begin{definition}
\label{def:drift_inducing:set}
A set of features $S$ is called \textdef{drift inducing} (a drift inducing set), iff $X_{S^C}$ is  not conditionally drifting given $X_S$. $S$ is \textdef{minimal drift inducing}, iff it is minimal with respect to inclusion, i.e.\ every proper subset $S' \subsetneqq S$ is not drift inducing.
\end{definition}

Unlike conditionally drifting sets of features, drift inducing sets are monotonic; extending a drift inducing set does not alter this property (see lemma~\ref{thm:extension_of_inducing_sets}). In this sense, it reflects the idea of an explanation, that contains all necessary, but maybe even more, causes for a certain event.  Moreover, monotonicity facilitates efficient algorithmic solutions for the problem.
Obviously, the set of all features explains any observed drift.  Hence it is reasonable to search for  minimal drift inducing sets.

 \begin{lemma}
 \label{thm:extension_of_inducing_sets}
 Drift inducing sets are monotonous, i.e., if $S$ is a drift inducing set, then every $S'$ with $S \subset S'$ is drift inducing. In particular, a feature, that is contained in every minimal drift inducing set, is contained in every drift inducing set.
 \end{lemma}


In a data analysis setup
we would like to figure out, which features were ``manipulated''\  in the first place and which features changed as a consequence of these manipulations.
Having this idea in mind, we consider a drift inducing set as promising candidate  for an explanation of which features induce the observed drift. Minimal drift inducing sets are local optima in the sense of the law of parsimony. 
First, we observe, that such optima actually exist: 

\begin{corollary}
Every drift inducing set contains a minimal drift inducing set. In particular, there exists at least one minimal drift inducing set for every  drift process.
\end{corollary}

So far, we have considered the idea of drift inducing feature sets. Now we investigate the question how to derive the relevance of single features based thereon, i.e.\
 we  derive a notion of drift inducing features. We have already seen, that inducing features must be contained in minimal drift inducing sets. It seems a reasonable strategy to assign more interest to features,  which are contained in every minimal set:

\begin{definition}
\label{def:drift_inducing:feature}
%

A feature is called \textdef{drift inducing} (a drift inducing feature), iff it is contained in some minimal drift inducing set. 
A drift inducing feature is called \textdef{strongly drift inducing}, iff it is contained in every drift inducing set, otherwise it is called \textdef{weakly drift inducing}.
%
\end{definition}

It is illustrative to compare the notion of drift inducing features with ideas of causal reasoning \cite{Schoelkopf}: In terms of causality, a latent variable $X_i$ is caused by some other latent variables $X_S$ with $S \subset R_i$, if we may find a function $f$, such that \begin{align}f(X_S,\varepsilon_i) = X_i,\end{align} where $\varepsilon_i$ is an independent noise variable. Typically, one assumes, that the noise is additive, i.e. $f$ factors as $f(X_S,\varepsilon_i) = g(X_S)+\varepsilon_i,$ while $\varepsilon_i$ is assumed to follow a known distribution, usually a normal distribution where $\varepsilon_i$ has small variance and $S$ is as small as possible such that it contains only relevant information.
%
Referring to conditional distributions yields to a comparable setting except for a lack of an explicit function $f$.
Instead of small noise, we are interested in the residual noise  ($\varepsilon_i$) being
non-drifting. 
The relation $f$ is only implicit, but time independent, hence the entire drift of $X_i$ must be induced by the drift of $X_S$.
Hence an inducing set $S$ is a set, which allows to give a non-drifting, causal, explanation for all features not contained in $S$, using $X_S$ as cause. In this sense,
%
drift inducing features are exactly those features, which are used for at least one 
non-drifting, minimal causal explanation, whereas strongly drift inducing features are needed for every non-drifting, causal explanation.

We can relate this notion to the simple feature wise notion of drift given in the last section: 

\begin{theorem}
\label{thm:inducing_and_drifting}
Drift inducing features are drifting. Furthermore, a feature is strongly drift inducing, if and only if it is strongly drifting. 
\end{theorem}

This theorem states, that if a feature is contained in any minimal drift inducing set, then it is drifting, too. On the other hand, it is not necessarily true, that a drifting feature is contained in any minimal drifting set, and therefore, may be ignored when we try to make sense of the ongoing drift. This observation allows us to characterise faithfully drifting features, which we described at the beginning of section~\ref{sec:setup}, by the following definition.

\begin{definition}
\label{def:faithfully}
A feature is called \textdef{faithfully drifting}, if it is drifting but not drift inducing.
\end{definition}

In the light of  causality, faithfully drifting features are those features, that can  be explained by means of a non-drifting causal explanation.
Furthermore, theorem \ref{thm:inducing_and_drifting} implies, that strongly drift inducing features, which are always needed to explain the drift of other features, cannot be explained by the drift of any other feature – in particular not by any other strongly drift inducing feature.
As a further consequence of theorem~\ref{thm:inducing_and_drifting}, it is easy to see, that faithfully drifting features may only be weakly drifting, which is reasonable, since  strongly drifting features contain relevant information with regard to the observed drift.

\begin{corollary}
\label{thm:weakly_inducing_is_weakly_drifting}
Faithfully drifting features are weakly drifting. In particular, weakly drifting features are either faithfully drifting or weakly drift inducing.
\end{corollary}

So far, it is an open question, whether faithfully drifting features exist. In the next section (theorem~\ref{thm:well_suited}), we will see, that in most cases every weakly drifting feature is actually faithfully drifting. 

\subsection{Suitability of the Drifting-Feature-Analysis Problem}
\label{sec:well_suited}
We are interested in the question whether the identification of these different types of drifting features provides the relevant information to explain drift in practical settings. 
In the beginning of section~\ref{sec:setup}, we defined the drifting-feature-analysis problem as the classification of all features into one of three categories: non-drifting, faithfully drifting and drift inducing. Non-drifting features are (implicitly) defined by definition~\ref{def:featuredrift} and this definition matches our description, as features, that do not drift. Faithfully drifting features were defined as those drifting features, whose drift can be completely explained by other features, see definition~\ref{def:faithfully}. Drift inducing features, in the sense of the problem, were defined as those features, that induce the observed drift of all other features, and their drift cannot be  explained by other features. This definition matches the definition of strongly inducing features, in the sense of definition~\ref{def:drift_inducing:feature}, since only the drift of such features may never be explained by the drift of other features (see theorem~\ref{thm:inducing_and_drifting}). We, therefore, restrict ourselves to the case, where there are only strongly drift inducing features and no ambiguities in the explanation of drift exists, i.e.\ we define the following:

\begin{definition}
\label{def:well_suited}
We say, that the drifting-feature-analysis problem is \textdef{well suited}, iff every drift inducing feature is strongly drift inducing.
\end{definition}

As a consequence of theorem~\ref{thm:inducing_and_drifting}, this is equivalent to the statement, that every feature whose drift may be explained by the drift of other features is automatically completely induced by other features, i.e.\ it is faithfully drifting. Hence this criterion is  a uniqueness criterion regarding the explanation of drift:

\begin{corollary}
Every drift inducing feature is strongly drift inducing, if and only if every weakly drifting feature is faithfully drifting.
\end{corollary}

Hence faithfulness constitutes a necessary criterion to assure, that the drifting-feature-analysis problem has a unique solution.
We are interested in a simple criterion, which guarantees that the drifting-feature-analysis is applicable in practical situations. It can be shown, that it is a sufficient condition, that every possible event may happen with a positive probability. This is fulfilled by many probability distributions, e.g.\ the normal distribution:

\begin{theorem}
\label{thm:well_suited}
If $\P_X$ has a strictly positive density, then the drifting-feature-analysis problem is well suited.
\end{theorem}

Using this theorem, we see, that we may always make use of the drifting-feature-analysis (in the sense of definition~\ref{def:well_suited}) to understand the ongoing drift, i.e. perform a feature wise classification. 
Now we turn to the problem of an algorithm for the identification of such features.
To do so, we relate the drifting-feature-analysis to classical settings of feature relevance determination and feature selection, which constitute well investigated problems in the literature and which offer efficient algorithmic solutions. 

\subsection{Related Work for Explaining Drift}
Up to our knowledge, 
there does not exist previous work which directly aims for an analysis of causal features for observed drift,
and there do not exist formalizations which can be related to the drifting-feature-analysis problem as stated above.
The closest approach, which we where able to find, is the work \cite{DBLP:journals/corr/WebbLPG17}, which provides a (potentially conditional) feature wise analysis by means of speed of change.
However, \cite{DBLP:journals/corr/WebbLPG17} only provides a tool for visualization drift by means of rate of change. No deeper theoretical considerations nor any guarantees on the degree or nature of insight were provided. The work \cite{ijcai2017-187} provides a way to transfer learning and learning with drift in the context of graphical models. The latter is related to our formalization, as we will see in  section~\ref{sec:connection}.

\section{Relation to Feature Relevance Determination and Algorithmic Solution}
In this section, we give a short introduction to the existing theory of feature relevance determination encountered in the literature, first. Then, we investigate the connection to the drifting-feature-analysis problem, and, in the last subsection, we describe an algorithmic approach to solve the feature-drift-analysis problem.

\subsection{Feature Relevance Theory}

In this section, we will consider  a regression problem over $\R^d\to\R$. 
We consider pairs of random variables $(X,Y)$, corresponding to data and label, where we use the same notation regarding the features of $X$ as before. 
The link to drifting-feature-analysis will later be based   on an identification of   $Y$ and $T$.

We repeat the notion of relevance of a feature to the label variable $Y$ as given by \cite{Kohavi_and_John}. The idea is, that a feature $X_i$ is relevant, if it contains information that may help us to predict $Y$. 

\begin{definition}
\label{def:relevance}
A feature $X_i$ is \textdef{strongly relevant} to $Y$, iff $X_i$ and $Y$ are not independent given the remaining features $X_{R_i}$, i.e. 
\begin{align}
    Y \not\!\amalg X_i | X_{R_i}.
\end{align}
where the symbol $\amalg$ refers to statistical independence of the random variables.

It is \textdef{weakly relevant} to $Y$, if it is not strongly relevant, but can be made strongly relevant by removing other features, i.e.\ there exists a subset $R' \subset R_i$, for which 
it holds
\begin{align}
    Y \not\!\amalg X_i | X_{R'}.
\end{align}
A feature is \textdef{relevant}, if it is either strongly or weakly relevant. Otherwise, it is called \textdef{irrelevant}.
\end{definition}

As pointed out in \cite{InterpretationofLinearClassifiersbyMeansofFeatureRelevanceBounds}, the distinction between strong and weak relevance is inspired by the observation, that some features may carry redundant information regarding $Y$. As an example, consider the case, where two features are identical copies of each other, i.e. $X = (X_1,X_2,X_2)$. Supposing, that $Y$ can be predicted using $X_1+X_2$, then the first feature is clearly relevant. The other  features carry relevant information but they are redundant and only one of those is required. In the framework of definition~\ref{def:relevance}, the first feature is strongly relevant, while features two and three are weakly relevant.


In the context of feature relevance, two problems are of particular interest, when dealing with sets of features: the minimal-optimal problem and the all-relevant problem. 
The \textit{minimal-optimal} problem refers to the problem of finding a (smallest) set of features, that are relevant to the Bayesian classifier and contain all important information; hence adding further features does not improve the prediction accuracy for  $Y$. In \cite{PolyTime}, this problem is linked to soft classification models, and this problem is related to
the challenge to identify  Markov blankets \cite{probabilisticreasoning}. 
It can be shown, that under the assumption, that $\P_X$ has a strictly positive density, there exists exactly one Markov blanket, which is then referred to as a Markov boundary. 
In particular, it was shown, that the Markov boundary, if it exists, is equivalent to the set of strongly relevant features in this case \cite{PolyTime}. This gives rise to an algorithm in linear time with respect to the number of features.

The \textit{all-relevant} problem refers to the problem of identifying all features relevant to $Y$. It was shown, that this problem needs exhaustive search for general distributions \cite{PolyTime}. However, under assumptions on the distribution (dubbed   as  PCWT-distribution), we can find an exact algorithm that solves this problem in quadratic time with respect to the number of features. 
Nonetheless, this algorithm relies on independence tests in every step, yielding a large computation time per step. Due to this restriction, other model based approaches were designed to overcome this problem \cite{InterpretationofLinearClassifiersbyMeansofFeatureRelevanceBounds,squamish}. 

In the next section, we will relate  the minimal-optimal and the all-relevant feature selection problem to  the drifting-feature-analysis problem.

\subsection{Drifting-Feature-Analysis as a Feature Relevance Problem}
\label{sec:connection}

In section~\ref{sec:def_drift}, we considered the conditional distribution of $X$ given $T$. This idea allows us to consider drift as the dependency of data and time, as  already presented in \cite{driftingdata}. In our setup, this is expressed by the following theorem.
\begin{theorem}
\label{thm:drift_as_independency}
$X$ has drift, if and only if $T \not\!\amalg X$. Furthermore, $Y$ has conditional drift given $X$, if and only if $T \not\!\amalg Y | X$.
\end{theorem}
 Bayes' rule enables us to link the conditional distributions $T$ and $X$ as follows:
\begin{align}
    p_t(X = x) = \frac{\P_X(x)}{\P_T(t)} \P[T = t | X = x].
\end{align}
This observation links drift of a distribution to the posterior of time given  an observation. This allows us to address drifting features via algorithms, which detect feature relevance for a posterior distribution or the associated regression problem. More precisely, we find the following:

\begin{theorem}
\label{thm:equivalence_to_relevance}
Assume, that $\P_X$ has a strictly positive density. Then, the drifting-feature-analysis problem is well suited, and it holds the following:
a feature $X_i$ is strongly relevant for the prediction $\P[T = t | X = x]$ if and only if
$X_i$ is drift inducing; 
a feature $X_i$ is weakly relevant if and only if it is faithfully drifting; it is irrelevant if and only if it is non-drifting.
%
\end{theorem}

This result induces algorithmic approaches for the drifting-feature-analysis problem, since it enables us to rely on established
techniques for the determination of strongly or weakly
relevant features for posterior distributions or the associated deterministic regression, respectively, provided the signal to noise ratio is sufficiently small.
Note that this observation is also related to the notion of
feature redundancy for weakly relevant features \cite{YuLiu}.

%

\subsection{Implementation Details}
\label{sec:implement}

As shown in theorem~\ref{thm:equivalence_to_relevance}, the drifting-feature-analysis problem is equivalent to the problem of finding all strongly and weakly relevant features for the 
probability distribution $\P[T=t|X=x]$, under the assumption of positive density, solving the minimal-optimal and the all-relevant problem. To do so, we will apply two approaches with different strengths and weaknesses, as we will see in section~\ref{sec:exp}.

Our first approach uses the Recursive-Independence-Test (RIT) \cite{PolyTime} to solve the all-relevant problem. The idea is to consider $T,X$ as a graphical model, i.e. $T,X_1,..,X_d$ are the nodes in a graph and there exists an edge, if and only if the respective nodes are not independent. Then, the all-relevant problem can be solved, by identifying the connected component in the graph, which contains the node $T$, this corresponds to all drifting features.
We apply a conditional independence test, to distinguish between strongly and weakly drifting features in this subgraph.    
Algorithmically, we rely on the  Hilbert-Schmidt-independence-criterion test \cite{HSIC} which uses a covariance analysis in kernel spaces and 
fast conditional independence test \cite{FCI}
, which is based on comparing the performance of random forests on different permutations, and the kernel independence test \cite{HSIC}, which are based on a kernelized covariance estimation. 
We will refer to this approach as statistical DFA.

Statistical DFA is computationally expensive and it may become instable in the case of many features.
Hence 
we consider an additional method which is presented in \cite{squamish}.
This has the restriction, that it can be used only in the case
that the probability distribution $\P[T=t|X=x]$ can approximately be modelled by a regression task $X\to T$ with sufficient signal to noise ratio. In this case,
the idea is to use a Random Forest classifier or regressor and analyze the resulting model by means of feature relevance. We will refer to this method as Relevance Bounds. Default parameters where used.\footnote{The source code of both methods is available on GitHub -- \url{https://github.com/FabianHinder/drifting-features-analysis}.}

\section{Experiments}
\label{sec:exp}
We evaluate both approaches of the drifting-feature-analysis on two standard benchmark data sets. We also created several ground truth data sets for the problem, in order to examine the performance of our method.

\subsection{Data Sets for Drifting-Feature-Analysis with Ground Truth}
\begin{figure}[!bt]
    \centering
    \includegraphics[width=0.9\textwidth]{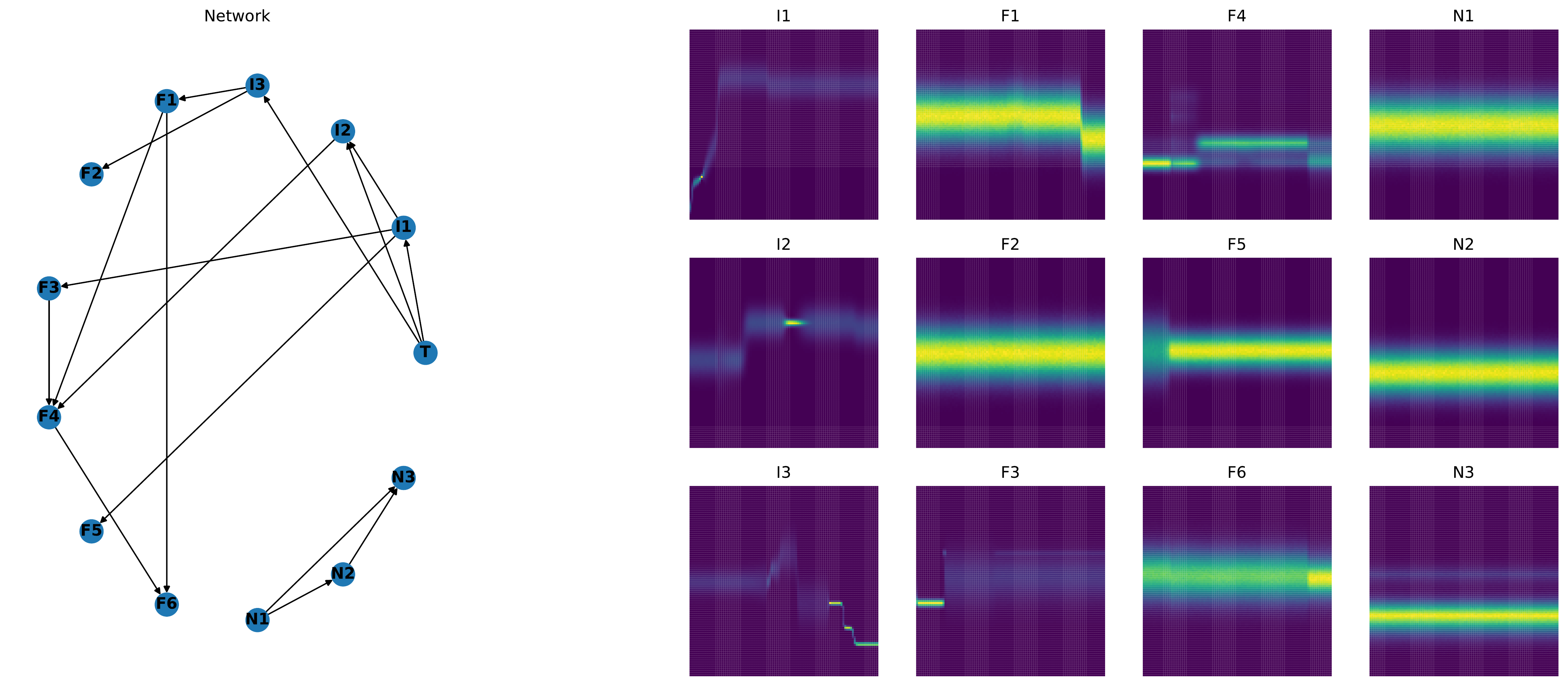}
    \caption{Visualization of an example network (left) for generating ground truth data and the distributions of its nodes plotted against time (right). The node names imply whether they are non-drifting (N), faithfully-drifting (F) or drift inducing (I).}
    \label{fig:ground_truth}
\end{figure}

We created a new benchmark data set with ground truth\footnote{The data set as well as the generating algorithms are available on GitHub -- \url{https://github.com/FabianHinder/drifting-features-analysis}.}, for the drifting-feature-analysis problem. It uses randomly constructed Bayesian networks. Besides the $X_i$-nodes, those networks also contain a $T$ node, as suggested in section~\ref{sec:def_drift}, which has no parent nodes. As implied by \cite{probabilisticreasoning}, and using theorem~\ref{thm:equivalence_to_relevance}, only the direct children of $T$ and their parents are drift inducing. All other nodes $X_j$ are either faithfully drifting or non-drifting, depending on whether there is a path between $X_j$ and $T$ or not.

%
%
We visualize an example network in figure~\ref{fig:ground_truth}. The nodes generate normally distributed samples; mean value and variance are computed using randomly initialized neural networks. Each data set generated by this method contains 10,000 samples.
\begin{wrapfigure}{r}{0.4\textwidth}
    \centering
    \includegraphics[width=0.4\textwidth]{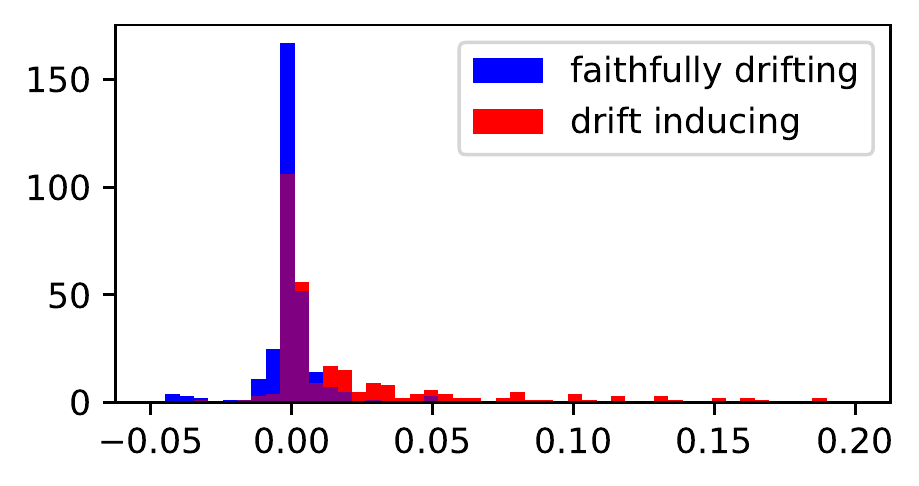}
    \caption{Leave-one-out-risk-estimation for random forests. Graphic shows $\hat{R}(X_{R_i})-\hat{R}(X)$ for 300 sample evaluations. }
    \label{fig:risk}
\end{wrapfigure}

\begin{table}[!b]
    \centering
    \caption{Results on data sets with known ground truth. The columns refer to the non-drifting (N), faithfully-drifting (F) or drift inducing (I) features and document the mean F1-Score. Score contains the mean micro-F1-Score, Accuracy refers to the accuracy of the internal model and Time is the mean run time in seconds. The standard deviation is below $0.05$ in all cases. }
    \begin{tabular}{|c|c|c||c|c|c|c|c|c||c|c|c|c|c|}
        \hline
         \multicolumn{3}{|c||}{Data set} & \multicolumn{6}{c||}{Relevance Bounds} & \multicolumn{5}{c|}{statistical DFA} \\
         I & F & N & I & F & N & Score & Accuracy & Time  & I & F & N & Score & Time \\
         \hline
5 & 15 & 5 & 	 $0.69$ &  $0.65$ &  $0.76$ &  $0.69$ & 0.98 & 20 &     $0.71$ &  $0.89$ &  $1.0$ &  $0.87$ & 127 \\    
7 & 13 & 5 & 	 $0.8$ &  $0.31$ &  $0.58$ &  $0.57$ & 0.88 & 20 &      $0.81$ &  $0.91$ &  $1.0$ &  $0.9$ & 128 \\    
6 & 11 & 8 & 	 $0.84$ &  $0.02$ &  $0.65$ &  $0.56$ & 0.90 & 18 &      $0.55$ &  $0.82$ &  $1.0$ &  $0.82$ & 111 \\    
6 & 14 & 5 & 	 $0.52$ &  $0.0$ &  $0.64$ &  $0.41$ & 0.85 & 20 &      $0.46$ &  $0.82$ &  $0.98$ &  $0.79$ & 126 \\    
6 & 8 & 11 & 	 $0.67$ &  $0.0$ &  $0.84$ &  $0.65$ & 0.93 & 17 &      $0.74$ &  $0.86$ &  $1.0$ &  $0.9$ & 92 \\    
6 & 7 & 12 & 	 $0.74$ &  $0.02$ &  $0.88$ &  $0.72$ & 0.96 & 20 &      $0.81$ &  $0.85$ &  $1.0$ &  $0.91$ & 86 \\    
         \hline
    \end{tabular}
    \label{tab:results:ground_truth}
\end{table}

We applied both approaches to several data sets, that were generated by the method described above. We report the mean category wise F1-score and the mean micro-F1-score over 10 runs. We also measured the mean run-time for each data set and method. The results are presented in table~\ref{tab:results:ground_truth}.
As can be seen, statistical DFA performs well on all data sets, in particular with respect to the detection of non-drifting features, but its run-time makes it inappropriate for real time application. Relevance Bounds, on the other hand, faces problems, which are caused by the limited accuracy of the underlying models, which correlates well with the overall score.

Both methods show some problems to distinguish between faithfully drifting and drift inducing features. In case of statistical DFA this is caused by the difficulty to predict causal directions from data, as discussed in \cite{DBLP:journals/jmlr/MooijPJZS16}. In case of Relevance Bounds this problem is caused by inaccurate risk-estimations, i.e.\ \cite{PolyTime} implies that if $\hat{R}(X_{R_i}) > \hat{R}(X) + \varepsilon$ then $i$ is drift inducing. We illustrated (the lack of) this behaviour in figure~\ref{fig:risk}.

\subsection{Benchmark Data Sets}
In addition to theoretical data with known ground truth, we applied our method to two, real world, benchmark data sets. 
The predictive value $T$ was created as  equidistant values, ranging from 0 to 1. 
We report the non-drifting, faithfully-drifting and drift inducing features for each data set and method. 

\paragraph{Poker}
In the Poker data set~\cite{poker}, each record is an example of a hand, consisting of five playing cards, drawn from a standard deck. Each card is described with suit and rank, resulting in 10 features.  1 mio  samples are sorted by the value of the hand, yielding strong drift. Both methods identify similar  features as faithfully drifting and drift inducing features, respectively: \\

\noindent
Relevance Bounds: \quad   N: [4, 6, 8], F: [7, 9], I: [0, 1, 2, 3, 5] \\
Statistic DFA:  \quad \quad \quad N: [2, 4, 6, 8], F: [7, 9], I: [0, 1, 3, 5] \\

\noindent
Note, that the cards are sorted in a hand according to the suit and rank, such that the suit and rank of the first cards limit the possible rank of the last cards of the hand, given the overall value  of the hand.
This fact is mirrored in the judgement of the rank of the last two cards as faithfully drifting features. Further, their color is not relevant for most hands, resulting in the judgement of the color of the last three (four) cards as irrelevant for the observed drift.

\paragraph{Electricity Market} 
The Electricity Market data set~\cite{electricitymarketdata} describes electricity pricing in South-East Australia. It records the price and demand in the states of New South Wales and Victoria as well as the amount of power transferred between those states. All time related features have been cleaned. Both methods classify the features in a similar way: \\

\noindent
Relevance Bounds: \quad   N: [-], F: [-], I: [0, 1, 2, 3] \\
Statistic DFA:  \quad \quad \quad N: [-], F: [1], I: [0, 2, 3] \\

\noindent
Since, there is no ground truth available in this case, we resort to adding two features, that are made up of Gaussian noise with different scales: \\

\noindent
Relevance Bounds: \quad   N: [4, 5], F: [-], I: [0, 1, 2, 3] \\
Statistic DFA:  \quad \quad \quad N: [4, 5], F: [1], I: [0, 2, 3] \\

\noindent
Both methods are able to identify these non drifting components.

%

\section{Discussion and Future Work}
In this paper we developed a method to explain feature wise drift by means of sets of drift inducing features. We developed a mathematical framework to formulate the problem of drifting-feature-analysis and compared it to notions of causality and feature relevance problems, deriving two detection methods this way, each with its own strengths and weaknesses.

We designed a method for generating ground truth data sets and evaluated the detection methods on  ground truth data and standard benchmark data sets. As it turned out, statistical DFA is yet computationally expensive. 
Relevance Bounds on the other hand is restricted to 
settings where the estimation of the posterior of time can be treated as a regression problem, yielding efficient estimations in this case.
It is subject of ongoing work to investigate density estimation and
associated feature relevance determination frameworks, instead.


\bibliographystyle{abbrv}
\bibliography{bib}

\newpage
\newenvironment{proof_}{\begin{proof}}{\end{proof}}

\addtocounter{definition}{-7}
\addtocounter{lemma}{-1}
\addtocounter{corollary}{-3}
\addtocounter{theorem}{-4}

\appendix

\section{Proofs}
In this section we provide proofs for the theorems and lemmas given in our paper. We also include all definitions, section headlines and the statements that are to be proven. The numbering is the same as in the paper.
 
 \subsection{Definition of Drift and Drifting Features}
 
\begin{definition}
The random variable $X$ has \textdef{drift} (or is drifting) iff
\begin{align}
    \P_T\times \P_T(\{(t,s)\in\T^2\:|\: p_t(X) &\neq p_s(X) \}) > 0.
\end{align}

We say, that the random variable $Y$ has \textdef{conditional drift} (or is conditionally drifting) given $X$ iff
\begin{align}
    \P_{T|X=x}\times\P_{T|X=x}(\{(t,s)\in\T^2\:|\: (p_t(Y|x) &\neq p_{s}(Y|x) \}) > 0
\end{align}
holds on a $\P_X$ non-null set.
\end{definition}

\begin{definition}
A feature $X_i$ is called \textdef{drifting}, iff there exists a set $R$, such that $X_i$ is conditionally drifting given $X_R$. It is called \textdef{strongly drifting} if we may choose $R = R_i$, otherwise it is called \textdef{weakly drifting}. 
\end{definition}

\subsection{Drift Inducing Features}

\begin{definition}
A set of features $S$ is called \textdef{drift inducing} (a drift inducing set), iff $X_{S^C}$ is  not  drifting conditionally given $X_S$. $S$ is \textdef{minimal drift inducing}, iff it is minimal with respect to inclusion, i.e.\ every proper subset $S' \subsetneqq S$ is not drift inducing.
\end{definition}
 
 \begin{lemma}
 Drift inducing sets are monotonous, i.e. if $S$ is a drift inducing set, then every $S'$ with $S \subset S'$ is drift inducing, too. In particular, a feature, that is contained in every minimal drift inducing set, is contained in every drift inducing set.
 \end{lemma}
 \begin{proof_}
 Using theorem~\ref{thm:drift_as_independency}, we may consider the problem in terms of statistical independence.
 We may write $S' = R \cup S$, with $R = S' \setminus S$. Using weak union, it follows, that
 \begin{align}
                    T \amalg X_{S^C} | X_S 
       &\Leftrightarrow T \amalg X_{S'^C}, X_R | X_S
     \\&\Rightarrow T \amalg X_{S'^C} | X_S, X_R
     \\&\Leftrightarrow T \amalg X_{S'^C} | X_{S'}.
 \end{align}
 \end{proof_}

\begin{corollary}
Every drift inducing set contains a minimal drift inducing set. In particular, there exists at least one minimal drift inducing set.
\end{corollary}
\begin{proof_}
Suppose, that there exists a drift inducing set $S$, which does not contain a minimal drift inducing set. In particular, $S$ is not minimal drift inducing itself, and therefore, cannot be the empty set. Since, $S$ is not minimal, it contains a proper subset $S' \subsetneqq S$, which is drift inducing, too. Since, every subset of $S'$ is a subset of $S$, it follows, that $S'$ does not contain a minimal drift inducing set, too. Since, $|S'| < |S| \leq d$, it follows by induction, that the empty set is drift inducing but not minimal drift inducing, which is a contradiction, since the empty set does not contain a proper subset.
\end{proof_}

\begin{definition}
A feature is called \textdef{drift inducing} (a drift inducing feature), iff it is contained in some minimal drift inducing set. 
A drift inducing feature is called \textdef{strongly drift inducing}, iff it is contained in every drift inducing set, otherwise it is called \textdef{weakly drift inducing}.
\end{definition}

\begin{theorem}
Drift inducing features are drifting. Furthermore, a feature is strong drift inducing, if and only if it is strong drifting. 
\end{theorem}
\begin{proof_}
Using theorem~\ref{thm:drift_as_independency}, we may consider the problem in terms of statistical independence. Let $X_i$ be a drift inducing feature. Hence, we may find an minimal drift inducing set $S$ that contains $i$, i.e. we may find $S$ such that $i \in S$, $T \amalg X_{S^C} | X_S$ and for every $S' \subsetneqq S$, in particular for $S' = S \setminus \{i\}$, we have ${T \not\!\amalg X_{S'^C} | X_{S'}}$. 
Now, suppose, that $X_i$ is not drifting. Then, for every $R \subset R_i$, we have, that ${T \amalg X_i | X_R}$. By setting $R = S \setminus \{i\}$ and applying contraction, we obtain
\begin{align}
    T \amalg X_{S^C} | X_R, X_i \wedge T \amalg X_i | X_R &\Rightarrow T \amalg X_{S^C}, X_i | X_R \Leftrightarrow T \amalg X_{R^C} | X_R. 
\end{align}
So $R \subsetneqq S$ is a drift inducing set. This is a contradiction. \\

Now, we show the second part:

\textbf{"$\Leftarrow$":} Let $X_i$ be a strong drifting feature. Suppose, there exists a minimal drift inducing set $S$ with $i \not\in S$. In particular, we have $S \subset R_i$. Then, by weak union, it follows
\begin{align}
    T \amalg X_{S^C} | X_S 
      &\Leftrightarrow T \amalg X_i,X_{S^C \setminus \{i\}} | X_S 
    \\&\Rightarrow T \amalg X_i | X_S,X_{S^C \setminus \{i\}}
    \\&\Leftrightarrow T \amalg X_i | X_{R_i}
\end{align}
which is a contradiction, since $X_i$ is strong drifting. Thus, $i$ has to be contained in every minimal drift inducing set, and therefore, we see, that $X_i$ is strong drift inducing.

\textbf{"$\Rightarrow$":} Let $X_i$ be a strong drift inducing feature. Suppose, that $X_i$ is not strong drifting. Then, it holds $T \amalg X_i | X_{R_i}$, and therefore, we see, that $R_i$ is a drift inducing set, which is a contradiction, since $i$ is contained in every drift inducing set (lemma~\ref{thm:extension_of_inducing_sets}) but $i \not\in R_i$.
\end{proof_}

\begin{definition}
A feature is called \textdef{faithfully drifting}, if it is drifting but not drift inducing.
\end{definition}

\begin{corollary}
Faithfully drifting features are weakly drifting. In particular, weakly drifting features are either faithfully drifting or weakly drift inducing.
\end{corollary}
\begin{proof_}
Suppose, that there exists a feature $X_i$, which is faithfully drifting and strong drifting, then, by theorem~\ref{thm:inducing_and_drifting}, it follows, that $X_i$ is strong drift inducing, and hence, drift inducing, which is a contradiction, since faithfully drifting features are not drift inducing.
\end{proof_}

\subsection{Suitability of the Drifting-Feature-Analysis Problem}

\begin{definition}
We say, that the drifting-feature-analysis problem is \textdef{well suited}, iff every drift inducing feature is strongly drift inducing.
\end{definition}

\begin{corollary}
Every drift inducing feature is strong drift inducing, if and only if every weakly drifting feature is faithfully drifting.
\end{corollary}
\begin{proof_}
Inducing features are either strong drift inducing or weakly drift inducing. By corollary~\ref{thm:weakly_inducing_is_weakly_drifting}, it follows, that weakly drifting features are either faithfully drifting or weakly drift inducing. If one of the statements above is true, then there are no weakly drift inducing features, and hence, the other follows.
\end{proof_}

\begin{theorem}
If $\P_X$ has a strictly positive density, then the drifting-feature-analysis problem is well suited.
\end{theorem}
\begin{proof_}
Using theorem~\ref{thm:drift_as_independency}, we may consider the problem in terms of independence. If $\P_X$ has a strictly positive density, then we may use the intersection property of conditional independence \cite[Theorem 25]{PolyTime} to show, that the intersection of two drift inducing sets is again drift inducing. Hence, it follows, that there exists exactly one minimal drift inducing set, so that every drift inducing feature is strong drift inducing.
\end{proof_}

\subsection{Feature Relevance Theory}

\begin{definition}
A feature $X_i$ is \textdef{strongly relevant} to $Y$, iff $X_i$ and $Y$ are not independent given the remaining features $X_{R_i}$, i.e. 
\begin{align}
    Y \not\!\amalg X_i | X_{R_i}.
\end{align}
where the symbol $\amalg$ refers to statistical independence of the random variables.

It is \textdef{weakly relevant} to $Y$, if it is not strongly relevant, but can be made strongly relevant by removing other features, i.e.\ there exists a subset $R' \subset R_i$, for which 
it holds
\begin{align}
    Y \not\!\amalg X_i | X_{R'}.
\end{align}
A feature is \textdef{relevant}, if it is either strongly or weakly relevant. Otherwise, it is called \textdef{irrelevant}.
\end{definition}

\subsection{Drifting Feature Analysis as a Feature Relevance Problem}

\begin{theorem}
$X$ has drift, iff $T \not\!\amalg X$. Furthermore, $Y$ has conditional drift given $X$, iff $T \not\!\amalg Y | X$.
\end{theorem}
\begin{proof_}
The first statement is proven in \cite{driftingdata}. For the second consider
\begin{align}
    0 &= \P_{X}[\P_{T|X=x}^2[p_t(Y|x) \neq p_s(Y|x)] > 0] \\
    \Leftrightarrow 1 &= \P_{X}(\{x\:|\:\P_{T|X=x}^2[p_t(Y|x) \neq p_s(Y|x)] > 0\}^C)
    \\&= \P_{X}[\P_{T|X=x}^2[p_t(Y|x) \neq p_s(Y|x)] = 0]
    \\&= \P_{X}[\P_{T|X=x}^2(\{(t,s)\:|\:p_t(Y|x) \neq p_s(Y|x)\}^C) = 1]
    \\&= \P_{X}[\P_{T|X=x}^2[p_t(Y|x) = p_s(Y|x)] = 1]
    \\&= \P_{X}[\P_{Y,T|X=x} = \P_{T|X=x} \times \P_{Y|X=x}] & \text{non-conditional case}
\end{align}
where we denote $\P_{T|X=x}^2 = \P_{T|X=x} \times \P_{T|X=x}$ the product measure of $\P_{T|X=x}$ with itself. 
Now, this is the case if and only if $Y$ and $T$ are conditionally independent given $X$.
\end{proof_}

\begin{theorem}
Assume, that $\P_X$ has a strictly positive density. Then, the drifting-feature-analysis problem is well suited, and it holds, that
a feature $X_i$ is strongly relevant, weakly relevant resp. irrelevant to the prediction of $T$, iff $X_i$ is drift inducing, faithfully drifting resp. non-drifting.
\end{theorem}
\begin{proof_}
Using theorem~\ref{thm:drift_as_independency}, it is easy to see, that 
the notions of strongly relevant, weakly relevant resp. irrelevant and strong drifting, weakly drifting resp. non-drifting, coincide. 
Now the statement follows by theorem~\ref{thm:well_suited}.
\end{proof_}

\end{document}